\documentclass{article}

\usepackage[final]{neurips_2021}
\usepackage[utf8]{inputenc} 
\usepackage[T1]{fontenc}    
\usepackage{verse}
\usepackage[dvipsnames]{xcolor}
\definecolor{darkpowderblue}{rgb}{0.0, 0.2, 0.6}
\usepackage[colorlinks=true,allcolors=darkpowderblue]{hyperref} 
\usepackage{url}            
\usepackage{microtype}      

\begin{document}

\title{BERTian Poetics:\\Constrained Composition with Masked LMs}

\author{%
Christopher Akiki and Martin Potthast\\
Text Mining and Retrieval Group\\
Leipzig University\\
Germany \\
\texttt{\{christopher.akiki,martin.potthast\}@uni-leipzig.de} \\
}

\maketitle

\begin{abstract}
    Masked language models have recently been interpreted as energy-based sequence models that can be generated from using a Metropolis--Hastings sampler. This short paper demonstrates how this can be instrumentalized for constrained composition and explores the poetics implied by such a usage. Our focus on constraints makes it especially apt to understand the generated text through the poetics of the OuLiPo movement.
\end{abstract}

\section{Introduction}
\label{sec:introduction}

The cultural capital \citep{bourdieu:2002} of poetry has rendered it fertile ground for computational approaches, making poetry generation ``an established research field'', recently surveyed by \citet{oliveira:2017}. This field spans a spectrum between two different perspectives which we find fitting to call, \textit{machine-as-poet} and \textit{user-as-poet}. The present work is aligned with the latter.

Machine-as-poet approaches encode \textit{poeticity} \citep{jakobson:1981} in a corpus and train a system to reproduce the prosodic and stylistic structure encoded within it \citep{ghazvininejad:2017,lau:2018}. User-as-poet approaches aim to produce tools that a \textit{poet-programmer} \citep{morris:2012a} or \textit{operator} \citep{james:2006} might use in their creative endeavors \citep{uthus:2019,zhipeng:2019}. Both approaches capitalize on recent advances in neural text generation, yet they exclusively involve auto-regressive language models.

The contribution of this paper is twofold: First, in Section~\ref{sec:methods}, we implement%
\footnote{\url{https://github.com/webis-de/ML4CD-21}}
the Metropolis--Hastings sampler introduced by \citet{goyal:2021} and use it to explore constrained composition using a masked language model.%
\footnote{Our implementation uses the base version of English roBERTa \citep{liu:2019} off the Hugging Face shelf \citep{wolf:2020}, but the ideas introduced here are agnostic to masked language model and language.}
Second, in Section~\ref{sec:poetics}, we reflect on useful vantage points through which to critically understand our approach, specifically how it might be enacting oulipian constraints \citep{symes:1999,james:2006}.
\section{Constrained Composition}
\label{sec:methods}

Transformers \citep{vaswani:2017} have allowed the scaling up of unconditional language models to unprecedent magnitudes \citep{brown:2020}. Such causal models can be used to generate text by defining a probability $P(\mathbf{X})$ over sequences that can be sampled from auto-regressively. In practice, however, they prove unwieldy \citep{holtzman:2020} and awkward to steer, rendering controllable generation a challenging open question \citep{weng:2021}, and out of reach for a flexible user-as-poet approach.

Instead, we turn our attention to the findings of \citet{goyal:2021}, who develop a tractable sampling scheme for masked language models. Even though they do not explicitly model sequences, masked language models are interpretable as energy-based sequence models that can be sampled from using the stationary distribution $P(\mathbf{X})$ of a Metropolis--Hastings Markov Chain.

In reality, the distribution being sampled from is $P(\mathbf{X} \mid l)$: a probability over all sequences of length~$l$. This constitutes the first constraint that the operator needs to set in advance and the only obligatory one. A prompting context $c$ can also be specified so that we sample from $P(\mathbf{X} \mid l, c)$. However, unlike auto-regressive models, this prompting context need not be causal but can span any subset of tokens. We refer to the general case of a non-contiguous context as \textbf{perforated prompting}. More generally, we introduce the following distribution over sequences and a list of poetic constraints $C_{1:n}$:
\begin{equation}
P(\mathbf{X}, C_{1:n}) = \prod_{i=1}^{n} P(C_i) \cdot P(\mathbf{X} \mid C_{1:n}),
\end{equation}
which can be read as a product of experts \citep{hinton:1999}, each constraining an aspect of the sequence. These constraints---be they syntactic,%
\footnote{e.g.: \textit{Lipogram}, a constraint which forbids the use of a given letter.}
semantic, or prosodic%
\footnote{e.g.: \textit{Bouts-Rim{\'e}s}, a literary game where a sequence of rhymes has to be expanded into a poem.}%
---can be enacted in logit space on an arbitrary subset of the sequence's tokens by setting the appropriate logits to $-\infty$ through any of the following operations at every sampling step (see Figure~\ref{fig:example_1} for examples):
\vspace{-1ex}
\begin{itemize}
\item
\textbf{Explicit prompting}: Restricts the vocabulary of the model to a chosen token.
\item
\textbf{Implicit prompting}: Restricts the vocabulary of the model to satisfy a constraint.
\end{itemize}

\begin{figure}[t]
\small
\begin{minipage}{0.5\linewidth}
\centering
\textbf{Beyond} those lines, the \textbf{unforeseen.}\\
The worst will never be predicted.\\
\textbf{Beyond} those lines, the \textbf{unforeseen.}\\
The unexpected will never be possible.
\end{minipage}%
\hfill%
\begin{minipage}{0.5\linewidth}
\raggedright
\textbf{I cannot} leave you. You have just passed away,\\
\textbf{I cannot} believe it! I have finally found freedom.
\end{minipage}
\caption{\footnotesize Explicit prompts in bold. \textbf{Left}: Lipogram through implicit prompting that filters out all vocabulary tokens containing the letter ``a''. \textbf{Right}: Prompting the masked language model with a left context.}
\label{fig:example_1}
\vspace{-2ex}
\end{figure}

\section{Poetics}
\label{sec:poetics}

Having defined all the constraints, as well as a seeding sequence,%
\footnote{The seeding sequence can also consist of \texttt{<mask>} tokens.}
the operator lets the sampling process run---potentially infinitely---which then proceeds to enumerate all token combinations of the model's vocabulary. This sampling process is modulated by knowledge gleaned from an immense training corpus, making certain combinations more likely than others.

The randomness inherent to sampling aligns this work with \textbf{aleatory poetics}, as the operator ``deliberately engages with chance as a compositional principle'' \citep{james:2012}, specifically as prompts make the starting text underdetermined by design, with the final text having a ``determinate final form'' \citeyearpar[ibid.][]{james:2012}. Another useful angle is what theorists refer to as \textbf{computational poetry}, whose defining feature is a procedure or algorithm that ``precedes and determines the poem'' \cite{morris:2012a} and sees the poet-programmer step back and attend to information flows \citeyearpar[ibid.][]{morris:2012a}. Given the central role of the machine in our poetic endeavor, the latter is also aligned with \textbf{conceptual poetry}, as the creation modalities are at least as important as the ideas they attempt to express \citep{perloff:2012a}. The universe that the generated text is allowed to inhabit and explore has to be predetermined by the operator, so that ``all of the planning and decisions are made beforehand and the execution is a perfunctory affair. The idea becomes a machine that makes the text'' \citep{goldsmith:2007}. 

Such programmatic principles notably modulate the work of the French \textbf{OuLiPo} movement, who use constraints to understand, ``explore and expand the field of literature'' \citep{poucel:2012}. In stark opposition to---and rejection of---the surrealist practice of automatic writing\footnote{This begs the facetious question, is GPT-3 \citep{brown:2020} a surrealist?} and the debilitating openness of contemporary writing, oulipian writing sees constraints as ``a generative tool that enables a conflict necessary for the renewal of poetic form'' \citep{deming:2009}. Such conflicts fuel a ``productive friction between the constraining algorithm and the author's desire for meaning'' \citep{james:2006}, allowing an artist to ``maximize their options through minimizing their choice'' \citep{symes:1999}.

\section{Ethics}
\label{sec:conclusion}
Concerns about the mechanistic aspects of the written word can be traced back to Plato's \textit{Phaedrus} \citep{plato:1972}. The field of neural text generation can only exponentiate these anxieties. Without due consideration and reflected usage, the large language models we have come to wield so readily can only calcify our existing biases, and potentially introduce others. The stochastic parrots introduced by \citet{bender:2021} stand to cause damage, through an intertextual pastiche of all that is ugly on the internet. This intertextual view of the generated text, alluded to but not named in \citet{bender:2021}, adds one final critical layer to our poetic enterprise, that of \textbf{found poetry} \citep{perloff:2012b}, except that the recontextualized collage of others' words is mediated within an impenetrable latent space through the weights of a neural network.

\bibliographystyle{chicago}
\bibliography{ml4cd21-bertian-poetics-lit}

\end{document}